\documentclass{article}
\usepackage{ijcai13}
\usepackage{pictex}
\usepackage{url}

\def\beq{\begin{equation}}
\def\eeq#1{\label{#1}\end{equation}}
\def\ba{\begin{array}}
\def\ea{\end{array}}
\def\i#1{\hbox{\it #1\/}}
\def\is#1{{\hbox{\scriptsize {\it #1\/}}}}

\def\optional#1{\empty}

\def\mvis{\!=\!}

\def\false{\hbox{\bf f}}
\def\true{\hbox{\bf t}}

\def\suf{\;\mbox{\large\boldmath $\Rightarrow$}\;}

\newtheorem{prop}{Proposition}

\newtheorem{lemma}{Lemma}

\def\proof{\noindent\i{Proof}.\hspace{3mm}}

\title{\bf Causal Laws and Multi-Valued Fluents}
 \author{ \ \\
 Enrico Giunchiglia, DIST --- Universit\`a di Genova\\
 Joohyung Lee and Vladimir Lifschitz, University of Texas at Austin\\
 Hudson Turner, University of Minnesota at Duluth}
\begin{document}

\bibliographystyle{named}

\date{}
\maketitle


\begin{abstract}
This paper continues the line of work on representing properties of
actions in nonmonotonic formalisms that stresses the distinction between being
\i{true} and being \i{caused}, as in the system of causal logic introduced
by McCain and Turner and in the action language $\cal C$ proposed by
Giunchiglia and Lifschitz.  The only fluents directly representable in
language $\cal C$ are truth-valued fluents, which is often inconvenient.
We show that both causal logic and language $\cal C$ can be extended to
allow values from arbitrary nonempty sets.  Our extension of
language $\cal C$, called ${\cal C}+$,  also makes it possible to describe
actions in terms of their attributes, which is important from the
perspective of elaboration tolerance.  We describe an embedding of~${\cal C}+$
in causal theories with multi-valued constants, relate ${\cal C}+$
to Pednault's action language ADL, and show how multi-valued constants can be
eliminated in favor of Boolean constants.
\end{abstract}

\section{Introduction}

This paper continues the line of work on representing properties of
actions in nonmonotonic formalisms that stresses the distinction between being
\i{true} and being \i{caused}.  Fangzhen Lin~\shortcite{lin95} extended the
situation calculus by atomic formulas $\i{Caused}(p,v,s)$ (fluent~$p$ is caused
to have value~$v$ in situation~$s$).  Causal theories in the sense of McCain
and Turner~\shortcite{mcc97} have postulates of the form $F\suf G$
(if~$F$ is true then~$G$ is caused).  The semantics
of the action language~$\cal C$~\cite{giu98} is based on the concept of a
``causally explained'' transition.  Solutions to the frame problem
expressed in these formalisms are applicable to domains in which actions
may have indirectly specified effects and qualifications.  This work has led
to the development of the Causal
Calculator\footnote{\tt http://www.cs.utexas.edu/users/tag/cc .}---an
automated system for planning and reasoning about actions in such domains.

The work described above is limited, however, to propositional (that is,
truth-valued)
fluents.  Lin's predicate \i{Caused} expects a propositional fluent as the
first argument and a truth value as the second.  Formulas $F$, $G$ in the
causal logic of McCain and Turner are propositional formulas.  Fluent symbols
in language $\cal C$ can represent propositional fluents only.  In applications
to representing specific domains, this is often inconvenient.  If we want to
express, for instance, that the current location of box~$B_1$ is~$L_2$, we
have to write $\i{Loc}(B_1,L_2)$ rather than $\i{Loc}(B_1)\mvis L_2$;
then the fact
that a box, when moved to another location, ``disappears'' from its
previous location can be treated as an indirect effect of the move action,
provided that appropriate causal laws are postulated.
In the notation of~$\cal C$, these laws are
\beq
\hbox{\ }\hspace{12mm}
{\bf caused}\ \neg\i{Loc}(b,l)\ {\bf if}\ \i{Loc}(b,l')
 \hspace{4mm}
  (l\neq l')\,.
\eeq{e0}
Similar causal laws are needed in many other cases.

In this paper we show how the system of causal logic from~\cite{mcc97}
and language $\cal C$ can be extended to allow values from
arbitrary nonempty sets, besides the set of Boolean truth values.  Our
extension of $\cal C$, denoted by ${\cal C}+$, makes it possible to represent
many action domains more concisely.  Language ${\cal C}+$ allows us also to
describe actions in terms of their attributes.  For instance, in~${\cal C}+$
we can express that robot $r$ moves box $b$ to location $l$ by
conjoining~$\i{Move}(b)$ with formulas~$\i{Mover}(b)\mvis r$
and~$\i{Destination}(b)\mvis l$, instead of using action names with many
arguments,
such as~$\i{Move}(r,b,l)$.  As argued in~\cite{lif00},
the use of action attributes can
help make a representation more elaboration tolerant in the sense
of~\cite{mcc99}.

As a preliminary step, in Section~\ref{sec:mvpl} we define the syntax and
semantics of propositional combinations of equalities; every equality contains
a nonlogical constant whose value may come from a certain nonempty domain.
Classical propositional logic corresponds precisely to the case when all
constants are Boolean (that is, have domain~${\{ \false,\true \}}$).
On the basis of that language, Section~\ref{sec:causal} defines causal theories
with multi-valued constants.  A simple characterization of the meaning of
``definite'' causal theories is provided by the concept of multi-valued
completion.
Section~\ref{sec:C+} introduces the action language~${\cal C}+$, shows an
example of its use, and defines a simple embedding of~${\cal C}+$ into
causal logic with multi-valued constants.
Section~\ref{sec:adl} relates ${\cal C}+$ to the language~ADL
from~\cite{ped94}.  Finally, Section~\ref{sec:elim}
describes methods of eliminating multi-valued constants
with finite domains in favor of Boolean constants.  The results of that
section show that, in the finite case, the extensions of causal logic and
language~$\cal C$ proposed in this paper can be reduced to the original
versions of these systems.

\section{Propositional Logic of Multi-Valued Constants}   \label{sec:mvpl}

A {\sl (multi-valued propositional) signature} is a set of symbols called
{\sl constants}, along with a nonempty set~$\i{Dom}(c)$ of symbols assigned
to each constant~$c$.  We call $\i{Dom}(c)$ the {\sl domain} of~$c$.  An
{\sl atom} of a signature~$\sigma$ is an expression of the form ${c\mvis v}$ 
(``the value of~$c$ is~$v$'') where $c \in \sigma$
and $v \in \i{Dom}(c)$.  A {\sl formula}
of~$\sigma$ is a propositional combination of atoms.   An
{\sl interpretation} of~$\sigma$ is a function that maps every
element of~$\sigma$ to an element of its domain.  An interpretation~$I$
{\sl satisfies} an atom ${c\mvis v}$ (symbolically, ${I\models c\mvis v}$)
if ${I(c)\mvis v}$.
The satisfaction relation is extended from atoms to arbitrary
formulas according to the usual truth tables for the propositional
connectives.  A {\sl model} of a set $X$ of formulas is an interpretation
that satisfies all formulas in $X$.  If every model of $X$ satisfies a
formula $F$ then we say that $X$ {\sl entails} $F$ and write $X\models F$.

A {\sl Boolean} constant is one whose domain is ${\{\false,\true\}}$.
When all constants are Boolean,
these definitions correspond to the usual syntax and semantics of
propositional formulas, if we agree to use only atoms of the
form~${c\mvis\true}$ and to use $c$ as shorthand for ${c\mvis\true}$.

We will often identify an interpretation~$I$ of a
multi-valued propositional signature with the set of atoms true in~$I$.

\section{Causal Theories with Multi-Valued Constants}  \label{sec:causal}

Causal theories are defined here in precisely the same manner as
in~\cite{mcc97}, except that formulas in the sense of the previous section
are allowed in place of propositional formulas.

\subsection{Syntax} \label{sec:causal-syntax}

Begin with a multi-valued propositional signature~$\sigma$.
By a {\sl causal law} we mean an expression of the form
\beq
  F \suf G
\eeq{eq:law}
where $F$ and $G$ are formulas of~$\sigma$.
By the {\sl antecedent} and {\sl consequent} of~(\ref{eq:law}),
we mean the formulas $F$ and $G$, respectively.  

Note that~(\ref{eq:law}) is not the material conditional 
$F \supset G$.
The intended reading of~(\ref{eq:law}) is:
\emph{Necessarily, if~$F$ then $G$~is caused.}

A {\sl causal theory} is a set of causal laws.

\subsection{Semantics} \label{sec:causal-semantics}

For any causal theory~$T$ and interpretation~$I$, let
$$
T^I =  \{\, G \,:\, \hbox{for some $F$, } F \suf G \in T 
                                    \hbox{ and } I \models F \,\}\,.
$$

So $T^I$ is the set of consequents of causal laws 
in~$T$ whose antecedents are true in~$I$.     
Intuitively then, $T^I$ entails exactly the formulas
that are caused to be true in~$I$ according to~$T$.  

An interpretation~$I$ is {\sl causally explained} according to
a causal theory~$T$ if 
$I$ is the unique model of~$T^I$.

It follows that
an interpretation~$I$ is causally explained according to~$T$ if and
only if, for every formula~$F$,
\begin{eqnarray*}
  I \models F & \hbox{iff} &T^I \models F\,.
\end{eqnarray*}
That is,
$I$~is causally explained according to~$T$ if and only if the
formulas that are true in~$I$ are exactly the formulas
caused to be true in~$I$ according to~$T$.
For further discussion of the so-called 
``principle of universal causation''
that motivates this definition, see \cite{mcc97,tur99}.

As an example, let $\sigma=\{c\}$ and $\i{Dom}(c)=\{1,2,\dots\}$, and let
the only causal law in $T$ be
\beq
c\mvis 1 \suf c\mvis 1\,.
\eeq{example}
The interpretation $I$ defined by $I(c)=1$ is causally explained according
to~$T$.  Indeed, $T^I=\{c\!\mvis\! 1\}$, so $I$ is the only model of~$T^I$.
Furthermore, $T$ has no other causally explained interpretations.
Indeed, for any interpretation $J$ such that $J(c)\neq 1$, $T^J$ is empty and
so has models different from~$J$.

\subsection{Multi-Valued Completion}

A causal theory $T$ is {\sl definite} if
\begin{itemize}
 \item no constant in the signature of~$T$ has a singleton domain,
 \item the consequent of every causal law of~$T$ is an atom or~$\bot$, and
 \item no atom is the consequent of infinitely many causal laws of~$T$.  
\end{itemize}

Due to the first two conditions,
an interpretation~$I$ is causally explained according to
a definite causal theory~$T$ if and only if ${I = T^I}$.

Definite causal theories have a concise translation into
sets of formulas, as follows.
For each atom~$A$ in the language of~$T$, the {\sl completion} of~$A$
is the formula
$$A \equiv \left(F_1 \vee \cdots \vee F_n\right)$$ 
where $F_1,\ldots,F_n$ ($n\geq 0$) are the antecedents of the causal laws 
with consequent~$A$.
The {\sl multi-valued completion} of~$T$ is obtained by taking
the completion of each atom in the language of~$T$, along with
the formula~$\neg F$ for each causal law
of the form~${F \suf \bot}$ that belongs to~$T$.

\begin{prop}  \label{completion}
Let $T$ be a definite causal theory.
The causally explained interpretations according to~$T$ are precisely
the models of the multi-valued completion of~$T$.
\end{prop}

For instance, causal theory (\ref{example}) is definite, and its multi-valued
completion is
$$
\ba{ll}
 \hspace{12mm} & c\mvis 1 \equiv c\mvis 1\,,\\
  & c\mvis v \equiv \bot \hspace{9mm} (v>1)\,.
\ea
$$
The interpretation causally explained according to~(\ref{example}) is the
only model of these formulas.

\optional{
\medskip\noindent\i{Proof of Proposition~\ref{completion}.\hspace{3mm}}
Assume that $I$ is causally explained according to~$T$.
Since $T$ is definite, it follows that ${I = T^I}$.
So for every atom~${c\mvis v}$ in~$I$, (i)~there is a formula~$F$
such that ${F \suf c\mvis v}$ belongs to~$T$ and ${I \models F}$, and
(ii)~there is no formula~$F$ and
${v' \in \i{Dom}(c)}$ different from~$v$ such that
${F \suf c\mvis v'}$ belongs to~$T$ and ${I \models F}$.
We can conclude that~$I$ satisfies the completion of every atom of~$\sigma$.
Similarly, since ${\bot \notin T^I}$, we know that
${I \models \neg F}$ for every 
causal law in~$T$ of the form~${F \suf \bot}$.
So $I$~is a model of the multi-valued completion of~$T$.
Proof in the other direction is similar.
}

\optional{
\vspace{2mm}
}

Multi-valued completion is a straightforward generalization of the
``literal completion'' method introduced in \cite{mcc97},
which resembles the well-known Clark completion method
for logic programming \cite{cla78}.  When all constants are Boolean,
multi-valued completion corresponds precisely to literal completion.

\section{Action Language ${\cal C}+$}  \label{sec:C+}

Action language ${\cal C}+$ is a multi-valued
extension of the action language~${\cal C}$ \cite{giu98},
and includes ${\cal C}$ as the special case in which
all constants are Boolean.

\subsection{Syntax}

Consider a multi-valued signature~$\sigma$ partitioned into
{\sl fluent symbols}~$\sigma^{\is{fl}}$
and {\sl action symbols}~$\sigma^{\is{act}}$.
A {\sl state formula} is a formula of signature~$\sigma^{\is{fl}}$.
An {\sl action} is an interpretation of signature~$\sigma^{\is{act}}$.

As an example, we will show how to use ${\cal C}+$ to describe several
boxes
that can be moved between various locations.  Take two collections of symbols,
\i{Boxes} and \i{Locations}.  For each $b \in \i{Boxes}$,
\begin{itemize}
\item
$\i{Loc}(b)$ is a fluent symbol with the domain $\i{Locations}$,
\item
$\i{Move}(b)$ is a Boolean action symbol, and
\item
$\i{Destination}(b)$ is an action symbol with the domain
$\i{Locations} \cup \{\i{None}\}$,
\end{itemize}
where $\i{None}$ is a symbol that does not belong to $\i{Locations}$.
To formalize the enhancement of this example mentioned in the introduction,
we would need also the action symbols~$\i{Mover}(b)$ with the domain
$\i{Robots} \cup \{\i{None}\}$.

There are two kinds of propositions in ${\cal C}+$: {\sl static laws}
of the form
\beq
{\bf caused}\ F\ {\bf if}\ G
\eeq{e-static}
and {\sl dynamic laws} of the form
\beq
{\bf caused}\ F\ {\bf if}\ G\ {\bf after}\ H\,,
\eeq{e-dynamic}
where $F$ and $G$ are state formulas
and $H$ is any formula of~$\sigma$.
In a proposition of either kind, formula~$F$ is called its {\sl head}.

For instance,
\beq
\ba c
{\bf caused}\ \bot\ {\bf if}\ \top
\qquad\qquad\qquad\qquad\qquad\qquad\qquad\\ \qquad\quad
     {\bf after}\ \i{Move}(b)\mvis\true\equiv\i{Destination}(b)\mvis\i{None}
\ea
\eeq{e1c}
is a dynamic law.  It expresses that $\i{Destination}(b)$ is an attribute
of action $\i{Move}(b)$: whenever a box $b$ is moved, it is moved to a certain
location, and the other way around.

An {\sl action description} is a set of propositions.

\subsection{Special Cases} \label{ss-special}

For any Boolean action symbol $\alpha$, state formula $F$ and formula $H$,
$$
\alpha\ {\bf causes}\ F\ {\bf if}\ H
$$
is shorthand for the dynamic law
$$
{\bf caused}\ F\ {\bf if}\ \top\ {\bf after}\ \alpha\mvis\true\wedge H\,.
$$
For instance,
\beq
\i{Move}(b)\ {\bf causes}\ \i{Loc}(b)\mvis l\ {\bf if}\ \i{Destination}(b)\mvis l
\eeq{e1d}
is a dynamic law.  Such expressions are used to describe direct effects of
actions.  More generally, a direct effect of the concurrent execution of
$\alpha_1,\dots,\alpha_k$
can be described by an expression of the form
$$
\alpha_1,\dots,\alpha_k\ {\bf causes}\ F\ {\bf if}\ H\,,
$$
which stands for
$$
{\bf caused}\ F\ {\bf if}\ \top\ {\bf after}\ 
   \alpha_1\mvis\true\wedge\cdots \wedge\alpha_k\mvis\true\wedge H\,.
$$

A dynamic law of the form
$$
\alpha_1,\dots,\alpha_k\ {\bf causes}\ \bot\ {\bf if}\ H
$$
can be further abbreviated as
$$
{\bf nonexecutable}\ \alpha_1,\dots,\alpha_k\ {\bf if}\ H\,.
$$
For instance,
\beq
\ba l
{\bf nonexecutable}\ \i{Move}(b)\
\qquad\qquad\qquad\qquad\\ \qquad\quad
 {\bf if}\ \i{Loc}(b)\mvis l \wedge \i{Destination}(b)\mvis l
\ea
\eeq{e1f}
is a dynamic law that asserts, intuitively, that it is impossible to
move $b$ to $l$ if $b$ is already at $l$.

For any state formula $F$,
$${\bf inertial}\ F$$
stands for the dynamic law
$$
{\bf caused}\ F\ {\bf if}\ F\ {\bf after}\ F\,.
$$
This proposition expresses that $F$
``tends to remain true,'' in the sense that if $F$ is true
before an action occurs then there will be a cause for $F$ if it
remains true after the action occurs.
For instance, the propositions
\beq
{\bf inertial}\ \i{Loc}(b)\mvis l
\eeq{e1g}
for all boxes $b$ and locations $l$ express that if a box does not change
its location then there is a cause for it to be where it is.
As observed by McCain and
Turner, such propositions can be used to express
the commonsense law of inertia.

For any state formula $F$, we write
$${\bf never}\ F$$
for the static law
$$
{\bf caused}\ \bot\ {\bf if}\ F\,.
$$
It represents a ``qualification
constraint'' in the sense of~\cite{lin94a}. For instance, the expressions
\beq
\hbox{\ }\hspace{15mm}
{\bf never}\ \i{Loc}(b)\mvis l \wedge \i{Loc}(b')\mvis l\hspace{3mm} (b\neq b')
\eeq{e1h}
are static laws; they allow us to conclude that two boxes cannot be
simultaneously moved to the same location, and that a box can be moved to a
location occupied by another box only if the latter is moved also.

\subsection{Semantics}

Consider an action description $D$.  A {\sl state} is an interpretation
of $\sigma^{\is{fl}}$ that
satisfies $G\supset F$ for every static law (\ref{e-static}) in $D$.
A {\sl transition}
is a triple $\langle s,a,s'\rangle$ where $s,s'$ are states
and $a$ is an action; $s$ is the {\sl initial} state of the transition, and
$s'$ is its {\sl resulting} state.
A formula $F$ is {\sl caused} in transition $\langle s,a,s'\rangle$
if it is
\begin{itemize}
\item the head of a static law (\ref{e-static}) from $D$ such that
${s' \models G}$, or
\item the head of a dynamic law (\ref{e-dynamic}) from $D$ such that
${s' \models G}$ and ${s\cup a \models H}$.
\end{itemize}
A transition $\langle s,a,s'\rangle$
is {\sl causally explained} according to $D$ if its resulting state $s'$
is the only interpretation of $\sigma^{\is{fl}}$ that satisfies all formulas
caused in this transition.

As with other action languages, we associate a
transition diagram to $D$:
the {\sl transition diagram} represented by $D$
is the labeled directed graph which has
the states of $D$ as nodes, and which includes an edge from $s$ to $s'$
labeled with action~$a$ if $\langle s,a,s'\rangle$ is a causally explained
transition of $D$.

\begin{figure}
\begin{center}
\input{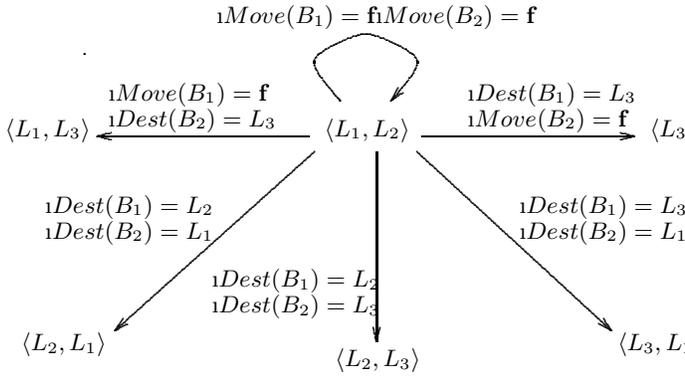}
\end{center}
\caption{A portion of the transition diagram for~(\ref{e1c})--(\ref{e1h})}
\label{figure}
\end{figure}

Consider the action description~(\ref{e1c})--(\ref{e1h})
with boxes~${B_1,B_2}$ and locations~${L_1,L_2,L_3}$. 
This system has six states, corresponding to the possible ways of placing
the two boxes in two of the three locations.  Each state is
reachable from any of the states by executing one action:
that is, for each pair of states $s,s'$ there is an
action~$a$ such that $\langle s,a,s'\rangle$ is causally explained according
to~(\ref{e1c})--(\ref{e1h}).  Figure~\ref{figure} shows the six
states, and the causally explained transitions $\langle s,a,s'\rangle$ in
which ${s(\i{Loc}(B_1))=L_1}$ and
${s(\i{Loc}(B_2))=L_2}$.  Each state~$s$ is represented
here by the pair of
locations~${\langle s(\i{Loc}(B_1)), s(\i{Loc}(B_2))\rangle}$.
Each action is represented by two of the four atoms true in it.

\subsection{Embedding in Causal Theories}

A ${\cal C}+$ action description can be translated as a causal theory.

Given a signature~$\sigma$ and number~$n\in\{0,1\}$,
define ${\sigma(n) = \{ c_n : c \in \sigma \}}$, with
${\i{Dom}(c_n) = \i{Dom}(c)}$.
For any formula~$F$ of~$\sigma$, let $F(n)$ be the formula
of~$\sigma(n)$
obtained by replacing each occurrence of each atom~${c\mvis v}$ in~$F$
by ${c_n\mvis v}$.  For any set~$T$ of formulas of~$\sigma$,
${T(n) = \{ F(n) : F \in T \}}$.  In particular, if $I$ is an interpretation
of $\sigma$ then $I(n)$ is an interpretation of~$\sigma(n)$.

For any action description~$D$, we obtain causal theory~$\i{ct}(D)$
over signature~${\sigma(0) \cup \sigma^\is{fl}(1)}$ as follows.
For each atom~$A$ of~$\sigma(0)$, include the causal law
\beq
  A \suf A\,.
\eeq{eq:exogenous-c}
For each static law~(\ref{e-static}) in~$D$, include the causal laws
\beq
  G(n) \suf F(n)
\eeq{eq:static}
for $n = 0,1$, and for each dynamic law~(\ref{e-dynamic}) in~$D$,
include the causal law 
\beq
  H(0) \wedge G(1) \suf F(1)\,.
\eeq{eq:dynamic}

For instance, if $D$ is action description (\ref{e1c})--(\ref{e1h}) 
then $\i{ct}(D)$ consists of the following causal laws:
$$
\begin{array}c
\i{Loc}_0(b)\mvis l \suf \i{Loc}_0(b)\mvis l\,, \\
\i{Move}_0(b)\mvis \false \suf \i{Move}_0(b)\mvis \false\,, \\
\i{Move}_0(b)\mvis \true \suf \i{Move}_0(b)\mvis \true\,, \\
\i{Destination}_0(b)\mvis v \suf \i{Destination}_0(b)\mvis v\,, \\
\i{Move}_0(b)\mvis\true\equiv\i{Destination}_0(b)\mvis\i{None}\suf\bot\,, \\
\i{Move}_0(b)\mvis \true \wedge\i{Destination}_0(b)\mvis l\suf \i{Loc}_1(b)\mvis l\,, \\
\i{Move}_0(b)\mvis \true \wedge \i{Loc}_0(b)\mvis l \wedge
                                 \i{Destination}_0(b)\mvis l \suf \bot\,,\\
\i{Loc}_0(b)\mvis l \wedge \i{Loc}_1(b)\mvis l \suf \i{Loc}_1(b)\mvis l\,,\\
\hspace{12mm}
\i{Loc}_n(b)\mvis l \wedge \i{Loc}_n(b')\mvis l \suf \bot \hspace{2mm}
  (b\!\neq\! b',\, n\!=\!0,\!1)\,.
\end{array}
$$

\begin{prop}  \label{embedding}
For any action description~$D$, a transition~${\langle s,a,s'\rangle}$
is causally explained according to~$D$ iff
the interpretation~${s(0) \cup a(0) \cup s'(1)}$ is causally
explained according to~$\i{ct}(D)$.  Moreover,
every interpretation causally explained according to~$\i{ct}(D)$
can be written in the form~${s(0) \cup a(0) \cup s'(1)}$
for some transition~${\langle s,a,s'\rangle}$.
\end{prop}

\optional{
\proof
We prove the second part first.  Assume that $I$ is
causally explained according to~$\i{ct}(D)$.
Clearly $I$ can be written in the form~${s(0) \cup a(0) \cup s'(1)}$,
where $s,s'$ are interpretations of~$\sigma^\is{fl}$ and
$a$ is an interpretation of~$\sigma^\is{act}$.  We need
to show that $s$ and $s'$ are states.  Consider any static
law~(\ref{e-static}) in~$D$.  Assume $s \models G$.
Then ${s(0) \models G(0)}$.  By~(\ref{eq:static}),
${G(0) \suf F(0) \in\i{ct}(D)}$, so
${F(0) \in \i{ct}(D)^I}$.  Since ${I \models \i{ct}(D)^I}$,
${s(0) \models F(0)}$. Hence ${s \models F}$.  We can conclude
that $s$ is a state.  The same argument shows that $s'$ is state.
}

\optional{
For the first part,
consider an arbitrary transition~${\langle s,a,s'\rangle}$
and let $I = {s(0) \cup a(0) \cup s'(1)}$.
Notice that
$\i{ct}(D)^I$ can be partitioned into three sets, as follows:
a set~$S_0$ of formulas of~$\sigma^\is{fl}(0)$,
a set~$A_0$ of formulas of~$\sigma^\is{act}(0)$, and
a set~$S_1$ of formulas of~$\sigma^\is{fl}(1)$.
Let $T$ be the set of formulas caused in~${\langle s,a,s'\rangle}$.  
Given the method for obtaining
causal laws~(\ref{eq:static}) and~(\ref{eq:dynamic}) from~$D$,
it is easy to verify that ${T(1) = S_1}$.  
}

\optional{
(Left-to-right)
Assume that ${\langle s,a,s'\rangle}$
is causally explained according to~$D$.
We show that $I$ is the unique model of~$\i{ct}(D)^I$
in three parts.
\begin{itemize}
 \item
Since $s'$ is the only interpretation of~$\sigma^\is{fl}$
that satisfies~$T$, $s'(1)$ is the only 
interpretation of~$\sigma^\is{fl}(1)$ that satisfies~${T(1) = S_1}$.
 \item
$A_0$ is obtained from the causal laws~(\ref{eq:exogenous-c})
for which~${A \in \sigma^\is{act}(0)}$.  In fact, these causal laws
yield exactly~$a(0)$, so $a(0)$~is the only interpretation
of~$\sigma^\is{act}$ that satisfies~$A_0$.
 \item
$S_0$ is obtained from
the causal laws~(\ref{eq:exogenous-c}) for which~${A \in \sigma^\is{fl}(0)}$,
and the causal laws~(\ref{eq:static}) for which~$n=0$.
In fact, the first family of causal laws contributes exactly~$s(0)$ to~$S_0$,
and since $s$~is a state, $s(0)$~satisfies the formulas obtained
from the second family of causal laws.  So $s(0)$~is the only
interpretation of~$\sigma^\is{fl}(0)$ that satisfies~$S_0$.
\end{itemize} 
}

\optional{
(Right-to-left)
Assume that $I$ is causally explained according to~$\i{ct}(D)$.
Since $I$ is the unique model of $\i{ct}(D)^I$, we know that
$s'(1)$ is the only 
interpretation of~$\sigma^\is{fl}(1)$ that satisfies~$S_1 = T(1)$.
Hence $s'$ is the only interpretation of~$\sigma^\is{fl}$
that satisfies~$T$, which shows that transition~${\langle s,a,s'\rangle}$
is causally explained according to~$D$.
}

\subsection{Definite Action Descriptions}

An action description $D$ is {\sl definite} if
\begin{itemize}
 \item no constant in the signature of $D$ has a singleton domain,
 \item the head of each proposition of $D$ is an atom or $\bot$, and
 \item no atom is the head of infinitely many propositions of $D$.
\end{itemize}

Notice that if an action description~$D$ is definite,
so is the corresponding causal theory~$\i{ct}(D)$.
Hence, by Propositions~\ref{completion} and \ref{embedding},
definite action descriptions have a concise translation into
multi-valued propositional logic.

For example, if $D$ is action description (\ref{e1c})--(\ref{e1h}) then
its translation into propositional logic with multi-valued constants is
equivalent to
$$
\begin{array}c
\i{Move}_0(b)\mvis\false\equiv\i{Destination}_0(b)\mvis\i{None}\,, \\
\neg (\i{Destination}_0(b)\mvis l\wedge \i{Loc}_0(b)\mvis l)\,, \\
\neg (\i{Loc}_n(b)\mvis l\wedge \i{Loc}_n(b')\mvis l)
\hspace{3mm}
(b\!\neq\! b',\, n\!=\!0,\!1)\,, \\
\i{Loc}_1(b)\mvis l\equiv 
\qquad\qquad\qquad\qquad\qquad\qquad\qquad\qquad\qquad\qquad\\ \hspace{1mm}
(\i{Destination}_0(b)\mvis l
\vee (\i{Loc}_0(b)\mvis l \wedge\i{Move}_0(b)\mvis \false)).
\end{array}
$$
These formulas entail, for instance, the two assertions mentioned at the
end of Section~\ref{ss-special}:
$$\begin{array} c
\neg(\i{Destination}_0(b)\mvis l \wedge \i{Destination}_0(b') \mvis  l)
 \hspace{3mm} (b\!\neq\!b')\,, \\
\i{Destination}_0(b) \mvis l \wedge \i{Loc}_0(b')\mvis l 
            \supset \i{Move}_0(b')\mvis \true\,.
\end{array}$$

\section{Relation of ${\cal C}+$ to ADL}\label{sec:adl}

Language ADL~\cite{ped94} describes effects of actions on multi-valued
fluents in terms of ``update conditions.''  We will show how this idea can
be incorporated in the syntactic framework of Section~\ref{sec:mvpl} and
then describe its relationship to language ${\cal C}+$.

As a preliminary step, consider a multi-valued propositional
signature~$\sigma$ whose constants have the same finite domain~\i{Dom}.
The concept of a formula of a signature $\sigma$
can be extended as follows.  A {\sl term} is a constant of $\sigma$, a value
(an element of \i{Dom}) or a variable (from a
fixed infinitely countable set).  An {\sl extended atom} is an expression of
the form $t\mvis v$ where $t$ is a term and $v$ is a value.  {\sl Extended
formulas} are formed from atoms using propositional connectives and
quantifiers, as in first-order logic.  We will sometimes identify a closed
extended formula $F$ with the formula in the sense of Section~\ref{sec:mvpl}
that is obtained from $F$ as follows: first, eliminate from $F$ all quantifiers
by replacing each subformula of the form $\forall x G(x)$ with
$\bigwedge_v G(v)$, where $v$ ranges over \i{Dom}, and each
$\exists x G(x)$ with $\bigvee_v G(v)$; second, replace all occurrences of
atoms of the form $v\mvis v$ with $\top$, and all occurrences of atoms of the
form $v_1\mvis v_2$, where $v_1$ is a value different from $v_2$, with~$\bot$.
This convention allows us, for instance, to talk about the satisfaction
relation between interpretations and closed extended formulas.

\subsection{ADL Action Descriptions}

Consider a multi-valued signature~$\sigma$ partitioned into
{\sl fluent symbols}~$\sigma^{\is{fl}}$
and {\sl action symbols}~$\sigma^{\is{act}}$, such that all fluent symbols
have the same finite domain, and all action symbols are Boolean.  An {\sl ADL
action description} consists of
\begin{itemize}
\item a closed extended formula $\i{Precond}^\alpha$ of signature
$\sigma^{\is{fl}}$ for every action symbol $\alpha$,
and
\item an extended formula $\i{Update}^\alpha_c(x)$ of signature
$\sigma^{\is{fl}}$, with no free variables other than $x$, for every action
symbol $\alpha$ and every fluent symbol $c$.
\end{itemize}
An  ADL action description is {\sl consistent} if, for every action
symbol~$\alpha$,
every fluent symbol $c$, and every pair of distinct values $v_1$, $v_2$,
$$\i{Precond}^\alpha\models
  \neg(\i{Update}^\alpha_c(v_1)\wedge\i{Update}^\alpha_c(v_2)).$$

Let $D$ be a consistent ADL action description, let $s$ and $s'$ be
interpretations of~$\sigma^\is{fl}$, and let $\alpha$ be an
action symbol.  We say that $s'$ is the {\sl result of executing $\alpha$
in $s$ according to $D$} if
$$s\models \i{Precond}^\alpha$$
and, for every fluent symbol $c$,
$$s'(c)=\cases{v,    &if $s\models \i{Update}^\alpha_c(v)$,\cr
               s(c), &if $s\models \neg\exists x\i{Update}^\alpha_c(x)$.
        }
$$

\subsection{Reduction to ${\cal C}+$}

The counterpart of an ADL action description~$D$ in language~${\cal C}+$
consists of the propositions
\beq
\ba l
{\bf inertial}\ c\mvis v\\
{\bf nonexecutable}\ \alpha\ {\bf if}\ \neg\i{Precond}^\alpha\\
\alpha\ {\bf causes}\ c \mvis v\ {\bf if}\ \i{Update}^\alpha_c(v)
\ea
\eeq{adl-translation}
for every fluent symbol $c$, value $v$ and action symbol $\alpha$.

In the following theorem, we identify a Boolean action symbol $\alpha$
with the action that maps $\alpha$ to $\true$ and maps every other action
symbol to $\false$.

\begin{prop} \label{adl}
For any consistent ADL action description $D$, $s'$ is the result of
executing $\alpha$ in $s$ according to $D$ iff
transition~$\langle s,\alpha,s'\rangle$ is causally explained according
to the counterpart of~$D$ in language ${\cal C}+$.
\end{prop}

We see that the version of ADL described above is significantly less
expressive than ${\cal C}+$:  ADL is mapped here into the
subset of ${\cal C}+$ that does not include static laws (and consequently
does not address the ramification problem) and has no concurrent actions or
noninertial fluents.

\optional{
\medskip\noindent\i{Proof of Proposition~\ref{adl}.\hspace{3mm}}
Recall that~(\ref{adl-translation}) stands for
$$
\ba l
{\bf caused}\ c\mvis v\ {\bf if}\ c\mvis v\  {\bf after}\ c\mvis v\\
{\bf caused}\ \bot\ {\bf if}\ \top\ 
    {\bf after}\ \alpha\mvis\true \wedge \neg\i{Precond}^\alpha\\
{\bf caused}\ c\mvis v\ {\bf if}\ \top\ 
    {\bf after}\ \alpha\mvis\true \wedge \i{Update}^\alpha_c(v).
\ea
$$
}

\optional{
(Left-to-right)
Assume that $s'$ is the result of executing~$\alpha$ in~$s$ according to $D$.
A formula is caused
in transition ${\langle s,\alpha,s'\rangle}$ iff it has the form $c=s'(c)$
for some fluent symbol $c$ (consider two cases, depending on whether
$s\models \i{Update}^\alpha_c(v)$ for some~$v$).  Consequently, this transition
is causally explained according to~(\ref{adl-translation}).
}

\optional{
(Right-to-left)
Assume that ${\langle s,\alpha,s'\rangle}$ is causally explained according
to~(\ref{adl-translation}).   Then $s\models\i{Precond}^\alpha$,
because otherwise~$\bot$ would be caused in this transition.
Take a fluent symbol $c$.  If, for some $v$,
$s\models \i{Update}^\alpha_c(v)$, then $c\mvis v$ is caused
in ${\langle s,\alpha,s'\rangle}$, so that $s'(c)=v$. 
If, on the other hand, $s\models \neg\exists x\i{Update}^\alpha_c(x)$
then $s'(c)=s(c)$.  Indeed, assume that $s'(c)\neq s(c)$.  Then no
formula of the form $c\mvis v$ is caused in this transition.  But
$|\i{Dom}| \ge 2$, because $s(c),s'(c)\in\i{Dom}$.  Consequently,
$s'$~cannot be the only
interpretation satisfying all caused formulas.
}

\section{Replacing Multi-Valued Constants with Boolean Constants}
                                                        \label{sec:elim}

If the domain of a constant~$c$ is finite, we can
replace~$c$ with a family of Boolean constants,
one for each element of~$\i{Dom}(c)$.
We first observe that this is easily done in propositional logic with
multi-valued constants.
This suggests a general method applicable to causal theories.
We then introduce a less obvious method that can be used to
preserve definiteness.
Finally, we specify similar elimination methods
for action language~${\cal C}+$.

\subsection{Eliminating Multi-Valued Constants from Formulas}

Let $\sigma$ be a multi-valued propositional signature, and let $c\in\sigma$.
Let $\sigma_c$ be the signature obtained from~$\sigma$
by replacing constant~$c$ with Boolean constants~$c(v)$
for all ${v \in \i{Dom}(c)}$.  For any formula~$F$ of $\sigma$,
let $F_c$ be the formula obtained by replacing each occurrence
of each atom ${c\mvis v}$ with ${c(v)\mvis\true}$.
For each interpretation~$I$ of~$\sigma$ there is
a corresponding interpretation~$I_c$ of~$\sigma_c$ such that
for all atoms~$A$ common to both signatures
\begin{eqnarray*}
I \models A & \hbox{iff} & I_c \models A\,,
\end{eqnarray*}
and for all $v \in \i{Dom}(c)$
\begin{eqnarray*}
  I \models c\mvis v & \hbox{iff} &
  I_c \models c(v)\mvis\true\,.
\end{eqnarray*}
The following lemma is easily proved by structural induction.

\begin{lemma}   \label{l:elim}
For any formula $F$, interpretation~$I$, and constant~$c$ of~$\sigma$,
${I \models F}$ iff ${I_c \models F_c}$\,.
\end{lemma}

Let $X$ be a set of formulas of signature~$\sigma$.
We'll say that a set~$X_c$ of formulas of signature~$\sigma_c$
{\sl correctly reduces} $c$ to a family of Boolean constants
(relative to~$X$)  if the following holds: an interpretation
of $\sigma_c$ is a model of~$X_c$ iff it corresponds to a model of~$X$.

Assuming that $\i{Dom}(c)$ is finite, let $\i{elim}_c$ be the formula
\beq
 \bigvee_v c(v)\mvis\true \;
             \wedge 
           \bigwedge_{v \neq v'} (c(v)\mvis\false \vee c(v')\mvis\false)\,.
\eeq{eq:elim}
The {\sl elimination} of~$c$ from~$X$
is the set of formulas of signature~$\sigma_c$ obtained by
replacing each occurrence of an atom~\hbox{$c\mvis v$}
in~$X$ with $c(v)\mvis\true$, and adding the formula~$\i{elim}_c$.

Notice that the models of~$\i{elim}_c$ are precisely the
interpretations of~$\sigma_c$ that correspond to an interpretation
of~$\sigma$.  This observation and Lemma~\ref{l:elim}
yield the following result.

\begin{prop}
Let $X$ be a set of formulas, and let~$c$ be a constant with a finite domain.
The elimination of~$c$ from~$X$ correctly reduces $c$
to a family of Boolean constants.
\end{prop}

\subsection{Eliminating Multi-Valued Constants from Causal Theories}
                                               \label{sec:elim-causal}

Begin with a causal theory~$T$ with signature~$\sigma$.
We'll say that a causal theory~$T_c$ with signature~$\sigma_c$
{\sl correctly reduces} $c$ to a family of Boolean constants
(relative to~$T$) if the following holds:
an interpretation of $\sigma_c$ is causally explained according to~$T_c$
iff it corresponds to an interpretation causally
explained according to~$T$.

\subsubsection{General Elimination Method for Causal Theories}

The {\sl general elimination} of~$c$ from~$T$
is the causal theory with signature~$\sigma_c$ obtained by
replacing each occurrence of an atom~$c\mvis v$
in~$T$ with $c(v)\mvis\true$, and adding the causal law
\beq
 \top \suf \i{elim}_c\,.
\eeq{eq:general-e}

\begin{prop}  \label{general}
Let $T$ be a causal theory with constant~$c$ whose domain is finite.
The general elimination of~$c$ from~$T$ correctly reduces $c$
to a family of Boolean constants.
\end{prop}

\optional{
\proof
Let $T_c$ be the general elimination of~$c$ from~$T$.
Because of~(\ref{eq:general-e}), any model of $T_c^I$, for
any interpretation~$I$ of~$\sigma_c$, corresponds to
an interpretation of~$\sigma$.
Consider any interpretations~$I,J$ of~$\sigma$, and
the corresponding interpretations~$I_c,J_c$ of~$\sigma_c$.
It is easy to verify (using Lemma~\ref{l:elim}) that
\begin{eqnarray*}
 I \models T^J & \hbox{iff} & I_c \models T_c^{J_c}\,.
\end{eqnarray*}
The result follows easily from these observations.
}

\optional{
\vspace{2mm}
}

This simple elimination method can be applied whenever $\i{Dom}(c)$ is finite,
but it does not preserve definiteness.  Since definiteness is quite useful,
we next introduce a less general elimination method that preserves it, and can
be applied to any definite causal theory in which $\i{Dom}(c)$ is finite.

\subsubsection{Definite Elimination Method for Causal Theories}

The {\sl definite elimination} of~$c$ from~$T$
is the causal theory with signature~$\sigma_c$ obtained by
replacing each occurrence of an atom~$c\mvis v$
in~$T$ with $c(v)\mvis\true$, and adding the causal laws
\beq
  c(v)\mvis\true \suf c(v')\mvis\false
\eeq{eq:definite1-e}
for all $v,v' \in \i{Dom}(c)$ such that $v \neq v'$,
and also adding
\beq
 \bigwedge_v c(v)\mvis\false \suf \bot\,.
\eeq{eq:definite2-e}

\begin{prop}    \label{definite}
Let $T$ be a causal theory with constant~$c$ such that
(i)~\i{Dom}(c) is finite, with at least two elements,
and (ii)~any consequent in which $c$ occurs is an atom.
The definite elimination of~$c$ from~$T$ correctly reduces $c$
to a family of Boolean constants.
\end{prop}

\optional{
We begin the proof of Proposition~\ref{definite}
with an easy consequence of Lemma~\ref{l:elim}.
}

\optional{
\begin{lemma}   \label{l:definite}
Let $T$ be a causal theory with constant~$c$, satisfying 
the conditions of Proposition~\ref{definite},
and let $T_c$ be the definite elimination of~$c$ from~$T$.
For any interpretations~$I,J$ of~$\sigma$, and
the corresponding interpretations~$I_c,J_c$ of~$\sigma_c$,
${I \models T^J}$ iff ${I_c \models T_c^{J_c}}$.
\end{lemma}
}

\optional{
\medskip\noindent\i{Proof of Proposition~\ref{definite}.\hspace{3mm}}
Let $T_c$ be the definite elimination of~$c$ from~$T$.
We must show that
an interpretation is causally explained
according to~$T_c$ iff it corresponds to an interpretation causally
explained according to~$T$.
}

\optional{
(Left-to-right)
For any interpretation~$I$ of~$\sigma_c$,
if $T_c^I$ is satisfiable,
then by~(\ref{eq:definite2-e})
at least one atom ${c(v)\mvis\true}$ belongs to~$I$, and
by~(\ref{eq:definite1-e}) at most one such atom belongs to~$I$.
It follows that any interpretation causally explained according
to~$T_c$ corresponds to an interpretation of~$\sigma$.  Given this,
the left-to-right direction follows easily from Lemma~\ref{l:definite}.
}

\optional{
(Right-to-left)
Assume that $I$ is causally explained according to~$T$.
So $I$ is the unique model of~$T^I$.
Since any formula of~$T^I$ in which $c$ occurs must be an atom,
and $\i{Dom}(c)$ has at least two elements,
we can conclude
that ${c\mvis v \in T^I}$, where $v$~is the element of~$\i{Dom}(c)$
such that ${I \models c\mvis v}$.  It follows by Lemma~\ref{l:elim}
that ${c(v)\mvis \true \in T_c^{I_c}}$.  Moreover, by~(\ref{eq:definite1-e}),
${c(v')\mvis \false \in T_c^{I_c}}$, for all ${v' \in \i{Dom}(c)}$
such that $v' \neq v$.  So any model of~$T_c^{I_c}$
corresponds to an interpretation of~$\sigma$.
Given this, it is straightforward (using Lemma~\ref{l:definite})
to show that $I_c$ is causally explained according to~$T_c$.
}

\optional{
\vspace{2mm}
}

\optional{
Perhaps we should say a word about the
requirement in Proposition~\ref{definite} that $|\i{Dom}(c)|$ exceed one.
For any ${c \in \sigma}$ such that ${\i{Dom}(c) = \{ v \}}$, ${I(c) = v}$ for
every interpretation~$I$ of~$\sigma$.  Accordingly,
if we wish to replace $c$ with a new Boolean
constant~$c(v)$, we can add the causal law~${\top \suf c(v)\mvis\true}$,
which is precisely what is done in the general elimination method.  By
comparison,
the definite elimination method would add only~${c(v)\mvis\false \suf \bot}$.
For example, take ${\sigma = \{c\}}$, ${\i{Dom}(c) = \{ v \}}$, and $T = \emptyset$.
Then $\sigma$ has only
one interpretation, and it is causally explained according to~$T$.
On the other hand, ${\sigma_c = \{c(v)\}}$,
${\i{Dom}(c(v)) = \{ \false,\true \}}$, and the
definite elimination of~$c$ from~$T$ is ${\{c(v)\mvis\false \suf \bot\}}$,
which has no causally explained interpretations.
}

\optional{
Recall that constants with singleton domains are also not allowed in definite causal
theories.  We now see that such constants are of little use.
Indeed, the preceding observations imply that if a causal theory~$T$
includes a constant~$c$ with ${\i{Dom}(c) = \{ v \}}$, then
every occurrence of atom~${c\mvis v}$ in~$T$ can be replaced with~$\top$
without affecting the causally explained interpretations.  In fact,
constant~$c$ can then be dropped from the signature~$\sigma$, and the
resulting causally explained interpretations will be precisely the
restrictions to~${\sigma \setminus \{c\}}$ of the interpretations
causally explained according to~$T$.
}

\subsection{Eliminating Multi-Valued Constants from ${\cal C}+$}
                                     \label{sec:elim-C+}

If an action description includes a constant~$c$
whose domain is finite,
$c$ can be replaced by
a family of Boolean constants using methods similar to
those introduced for causal theories.

\subsubsection{Eliminating Multi-Valued Fluent Constants}

Let $D$ be an action description with fluent symbol~$c$.
We'll say that an action description~$D_c$
with action symbols~$\sigma^\is{act}$
and fluent symbols~$\sigma_c^\is{fl}$
{\sl correctly reduces} $c$ to a family of Boolean fluent constants
(relative to~$D$) if the following hold.
\begin{itemize}
 \item If ${\langle s,a,s'\rangle}$ is causally explained according to~$D$,
  then ${\langle s_c,a,s_c'\rangle}$ is causally explained according to~$D_c$.
 \item Any transition causally explained according to~$D_c$ can be written
in the form ${\langle s_c,a,s_c'\rangle}$ where
${\langle s,a,s'\rangle}$ is causally explained according to~$D$.
\end{itemize}

\subsubsection{General Elimination for Fluents in Action Descriptions}

The {\sl general elimination} of fluent symbol~$c$ from~$D$
is the action description
with action symbols~$\sigma^\is{act}$
and fluent symbols~$\sigma_c^\is{fl}$ obtained by
replacing each occurrence of an atom~${c\mvis v}$
in~$D$ with $c(v)\mvis\true$, and adding the static law
\beq
{\bf caused}\ \i{elim}_c\ {\bf if}\ \top\,.
\eeq{C-general-e}

\begin{prop}
Let $D$ be an action description with fluent symbol~$c$
whose domain is finite.
The general elimination of~$c$ from~$D$ correctly reduces $c$
to a family of Boolean fluent constants.
\end{prop}

\optional{
\proof
Let $D_c$ be the general elimination of~$c$ from~$D$.
Because of~(\ref{C-general-e}), any state of~$D_c$
corresponds to an interpretation of~$\sigma^\is{fl}$.
Moreover, by Lemma~\ref{l:elim} it follows that
the states of~$D_c$ correspond precisely to the states of~$D$.
So any transition of~$D_c$ can be written
in the form ${\langle s_c,a,s_c'\rangle}$ where
${\langle s,a,s'\rangle}$ is a transition of~$D$.
It is easy to verify (again using Lemma~\ref{l:elim}) that
${\langle s,a,s'\rangle}$ is causally explained according
to~$D$ iff ${\langle s_c,a,s_c'\rangle}$ is causally explained according
to~$D_c$.
}

\subsubsection{Definite Elimination for Fluents in Action Descriptions}

The {\sl definite elimination} of fluent symbol~$c$ from~$D$
is the action description
with action symbols~$\sigma^\is{act}$
and fluent symbols~$\sigma_c^\is{fl}$ obtained by
replacing each occurrence of an atom~${c\mvis v}$
in~$D$ with $c(v)\mvis\true$, and adding the static laws
\beq
{\bf caused}\ 
   c(v)\mvis\false\ {\bf if}\ c(v')\mvis\true
\eeq{eq:definite1-ce}
for all $v,v' \in \i{Dom}(c)$ such that $v \neq v'$, and also
adding the static law
\beq
{\bf caused}\ \bot\ {\bf if}\ 
  \bigwedge_v c(v)\mvis\false\,.
\eeq{eq:definite2-ce}

Note that the static laws~(\ref{eq:definite1-ce}) generalize
example~(\ref{e0}) from the introduction.

\begin{prop}
Let $D$ be an action description with fluent symbol~$c$
such that
(i)~\i{Dom}(c) is finite, with at least two elements,
and (ii)~any head in which $c$ occurs is an atom.
The definite elimination of~$c$ from~$D$ correctly reduces
$c$ to a family of Boolean fluent constants.
\end{prop}

\optional{
\proof
Let $D_c$ be the definite elimination of~$c$ from~$D$.
Because of~(\ref{eq:definite1-ce}) and (\ref{eq:definite2-ce}),
any state of~$D_c$
corresponds to an interpretation of~$\sigma^\is{fl}$.
By Lemma~\ref{l:elim}, it follows that
the states of~$D_c$ correspond precisely to the states of~$D$.
So any transition of~$D_c$ can be written
in the form ${\langle s_c,a,s_c'\rangle}$ where
${\langle s,a,s'\rangle}$ is a transition of~$D$.
It remains to show that
${\langle s,a,s'\rangle}$ is causally explained according
to~$D$ iff ${\langle s_c,a,s_c'\rangle}$ is causally explained according
to~$D_c$.
}

\optional{
Consider any transition ${\langle s,a,s'\rangle}$ of~$D$.
By Proposition~\ref{embedding},
${\langle s,a,s'\rangle}$ is causally explained according
to~$D$ iff ${s(0) \cup a(0) \cup s'(1)}$ is causally explained according
to~$\i{ct}(D)$.  Let $T$ be the definite elimination of~$c_1$ from~$\i{ct}(D)$.
By Proposition~\ref{definite},
${s(0) \cup a(0) \cup s'(1)}$ is causally explained according to~$\i{ct}(D)$ iff
${I = s(0) \cup a(0) \cup s'(1)_{c_1}}$ is causally explained according to~$T$.
Let $X$ be the set of all formulas over~$\sigma(0)$.
By considering the form of~$T$, one can show the following.
\begin{itemize}
 \item ${I \cap X \subseteq T^I}$.
 \item Since $s$ is a state, ${I \models T^I \cap X}$.
 \item Every formula in ${T^I \setminus X}$
is a formula of~$\sigma^\is{fl}(1)_{c_1}$.
\end{itemize}
So $I$ is causally explained according to~$T$ iff ${I \setminus X}$~is
the unique model of~${T^I \setminus X}$.
We complete the proof by showing that
${I \setminus X}$~is
the unique model of~${T^I \setminus X}$ iff
${\langle s_c,a,s_c'\rangle}$ is causally explained according
to~$D_c$.
}

\optional{
It is enough to show that for
any formula~$F$ of $\sigma^\is{fl}$, ${F(1)_{c_1} \in T^I}$
iff $F_c$~is caused in~${\langle s_c,a,s_c'\rangle}$ (according
to~$D_c$).  So let $F$ be a formula of~$\sigma^\is{fl}$.
We need to consider each proposition in~$D_c$ whose head is~$F_c$;
these correspond exactly to the causal laws in~$T$
whose consequent is~$F(1)_{c_1}$.
\begin{itemize}
 \item There is a static law
   $${{\bf caused}\ F_c\ {\bf if}\ G_c \in D_c}$$ iff
   $${G(1)_{c_1} \suf F(1)_{c_1} \in T}\,.$$
  And since
  \begin{eqnarray*}
    s_c' \models G_c & \hbox{iff} & s'(1)_{c_1} \models G(1)_{c_1}\,,
  \end{eqnarray*}
  no problem here.
 \item  There is a dynamic law
   $${{\bf caused}\ F_c\ {\bf if}\ G_c\ {\bf after} H_c \in D_c}$$ iff
   $${G(1)_{c_1} \wedge H(0) \suf F(1)_{c_1} \in T}\,.$$
   Here we notice again that
  \begin{eqnarray*}
    s_c' \models G_c & \hbox{iff} & s'(1)_{c_1} \models G(1)_{c_1}\,,
  \end{eqnarray*}
   and also observe that
   $$\begin{array}{rcl}
    \hspace{4mm}
    s_c \cup a \models H_c & \hbox{iff} & (s \cup a)_c \models H_c  \\
     & \hbox{iff} & s \cup a \models H \hspace{6mm} (\hbox{by Lemma 1}) \\
     & \hbox{iff} & s(0) \cup a(0) \models H(0)\,.
   \end{array}$$
\end{itemize}
}

\subsubsection{Eliminating Multi-Valued Action Constants}

Let $D$ be an action description with action symbol~$c$.
We'll say that an action description~$D_c$
with action symbols~$\sigma_c^\is{act}$
and fluent symbols~$\sigma^\is{fl}$
{\sl correctly reduces} $c$ to a family of Boolean action constants
(relative to~$D$) if the following hold.
\begin{itemize}
 \item If ${\langle s,a,s'\rangle}$ is causally explained according to~$D$,
  then ${\langle s,a_c,s'\rangle}$ is causally explained according to~$D_c$.
 \item Any transition causally explained according to~$D_c$ can be written
in the form ${\langle s,a_c,s'\rangle}$ where
${\langle s,a,s'\rangle}$ is causally explained according to~$D$.
\end{itemize}

The {\sl elimination} of action symbol~$c$ from~$D$
is the action description
with action symbols~$\sigma_c^\is{act}$
and fluent symbols~$\sigma^\is{fl}$ obtained by
replacing each occurrence of an atom~$c\mvis v$
in~$D$ with $c(v)\mvis\true$, and adding the dynamic law
\beq
{\bf caused}\ \bot\ {\bf if}\ \top\ {\bf after}\ \neg\i{elim}_c\,.
\eeq{C-actions-e}

\begin{prop}
Let $D$ be an action description with action symbol~$c$
whose domain is finite.
The elimination of~$c$ from~$D$ correctly reduces $c$
to a family of Boolean action constants.
\end{prop}

\optional{
\proof
Let $D_c$ be the elimination of action symbol~$c$ from~$D$.
By (\ref{C-actions-e}), 
any transition causally explained according to~$D_c$ can be written
in the form ${\langle s,a_c,s'\rangle}$ where
${\langle s,a,s'\rangle}$ is a transition of~$D$.
The result then follows easily from the observation that
(by Lemma~\ref{l:elim})
the formulas caused in ${\langle s,a,s'\rangle}$ (according to~$D$)
are exactly the formulas caused in ${\langle s,a_c,s'\rangle}$
(according to~$D_c$).
}

\optional{
\subsubsection{Example}
}

\optional{
Consider the action description~(\ref{e1c})--(\ref{e1h}).
Let's eliminate all non-Boolean constants.
}

\optional{
For fluents $\i{Loc}(b)$ add propositions~(\ref{e0}),
as discussed in the introduction, and also
add the static law
$$
{\bf caused}\ \bot\ {\bf if}\ 
  \bigwedge_l \neg\i{Loc}(b,l)\,.
$$
}

\optional{
For actions $\i{Destination}(b)$, add
$$
{\bf caused}\ \bot\ {\bf if}\ \top\ {\bf after}\
   \bigwedge_v \neg \i{Destination}(b,v)$$
and, for all $v,v' \in \i{Dom}(\i{Destination}(b))$ such that $v \neq v'$,
also add
$${\bf nonexecutable}\ \i{Destination}(b,v), \i{Destination}(b,v')\ {\bf if}\ \top\,.
$$
(These propositions are obtained by simplifying~(\ref{C-actions-e}).)
}

\optional{
Finally, replacing non-Boolean constants in~(\ref{e1c})--(\ref{e1h}) yields
the following propositions.
$$
\ba c
{\bf caused}\ \bot\ {\bf if}\ \top\ 
     {\bf after}\ \i{Move}(b) \equiv\i{Destination}(b,\i{None}) \\
\i{Move}(b)\ {\bf causes}\ \i{Loc}(b,l)\ {\bf if}\ \i{Destination}(b,l) \\
{\bf nonexecutable}\ \i{Move}(b)\ 
 {\bf if}\ \i{Loc}(b,l) \wedge \i{Destination}(b,l) \\
{\bf inertial}\ \i{Loc}(b,l) \\
\hbox{\ }\hspace{15mm}
{\bf never}\ \i{Loc}(b,l) \wedge \i{Loc}(b',l) \hspace{3mm} (b\neq b')
\ea
$$
}

\section{Conclusion}

The extension of the action language~$\cal C$ proposed in this paper
is an improvement of that language in two ways.  First, we can now represent
multi-valued fluents directly, without replacing them by Boolean fluents.
Second, actions can be now described in terms of their attributes.
This work provides mathematical
background for adding these features to the input
language of the Causal Calculator, which will turn it into a better
knowledge representation tool.

\section{Acknowledgements}

We are grateful to Esra Erdem for comments on an earlier draft.
The first author was partially supported by ASI and MURST.  The second and
third authors were partially supported by NSF under grant IIS-9732744.
The fourth author was partially supported by NSF CAREER Grant \#0091773.


\begin{thebibliography}{}

\bibitem[\protect\citeauthoryear{Clark}{1978}]{cla78}
Keith Clark.
\newblock Negation as failure.
\newblock In Herve Gallaire and Jack Minker, editors, {\em Logic and Data
  Bases}, pages 293--322. Plenum Press, New York, 1978.

\bibitem[\protect\citeauthoryear{Giunchiglia and Lifschitz}{1998}]{giu98}
Enrico Giunchiglia and Vladimir Lifschitz.
\newblock An action language based on causal explanation: Preliminary report.
\newblock In {\em Proc.~AAAI-98}, pages 623--630. AAAI Press, 1998.

\bibitem[\protect\citeauthoryear{Lifschitz}{2000}]{lif00}
Vladimir Lifschitz.
\newblock Missionaries and cannibals in the causal calculator.
\newblock In {\em Principles of Knowledge Representation and Reasoning:
  Proc.~Seventh Int'l Conf.}, pages 85--96, 2000.

\bibitem[\protect\citeauthoryear{Lin and Reiter}{1994}]{lin94a}
Fangzhen Lin and Raymond Reiter.
\newblock State constraints revisited.
\newblock {\em Journal of Logic and Computation}, 4:655--678, 1994.

\bibitem[\protect\citeauthoryear{Lin}{1995}]{lin95}
Fangzhen Lin.
\newblock Embracing causality in specifying the indirect effects of actions.
\newblock In {\em Proc.~IJCAI-95}, pages 1985--1991, 1995.

\bibitem[\protect\citeauthoryear{McCain and Turner}{1997}]{mcc97}
Norman McCain and Hudson Turner.
\newblock Causal theories of action and change.
\newblock In {\em Proc.~AAAI-97}, pages 460--465, 1997.

\bibitem[\protect\citeauthoryear{McCarthy}{1999}]{mcc99}
John McCarthy.
\newblock Elaboration Tolerance.
\newblock In progress, 1999. Available at \url{http://www-formal.stanford.edu/jmc/elaboration.html}.

\bibitem[\protect\citeauthoryear{Pednault}{1994}]{ped94}
Edwin Pednault.
\newblock {ADL} and the state-transition model of action.
\newblock {\em Journal of Logic and Computation}, 4:467--512, 1994.

\bibitem[\protect\citeauthoryear{Turner}{1999}]{tur99}
Hudson Turner.
\newblock A logic of universal causation.
\newblock {\em Artificial Intelligence}, 113:87--123, 1999.

\end{thebibliography}

\end{document}